 \documentclass[tablecaption=bottom]{jmlr}\usepackage[]{graphicx}\usepackage[]{color}
\makeatletter
\def\maxwidth{ %
  \ifdim\Gin@nat@width>\linewidth
    \linewidth
  \else
    \Gin@nat@width
  \fi
}
\makeatother

\definecolor{fgcolor}{rgb}{0.345, 0.345, 0.345}

\usepackage{framed}
\makeatletter
 {\par\unskip\endMakeFramed%
 \at@end@of@kframe}
\makeatother

\definecolor{shadecolor}{rgb}{.97, .97, .97}
\definecolor{messagecolor}{rgb}{0, 0, 0}
\definecolor{warningcolor}{rgb}{1, 0, 1}
\definecolor{errorcolor}{rgb}{1, 0, 0}

\usepackage{alltt} 

\usepackage{booktabs}
\usepackage[load-configurations=version-1]{siunitx} 

\theorembodyfont{\upshape}
\theoremheaderfont{\scshape}
\theorempostheader{:}
\theoremsep{\newline}

\jmlrvolume{1}
\jmlryear{2018}
\jmlrsubmitted{23.06.2018}
\jmlrworkshop{ICML 2018 AutoML Workshop} 

\title{Automatic Gradient Boosting}

  \author{\Name{Janek Thomas} \Email{janek.thomas@stat.uni-muenchen.de}\\
   \Name{Stefan Coors} \Email{stefan.coors@campus.lmu.de}\\\Name{Bernd Bischl} \Email{bernd.bischl@stat.uni-muenchen.de}\\
   \addr Department of Statistics, LMU, Ludwigstrasse 33, D80539 Munich}

\IfFileExists{upquote.sty}{\usepackage{upquote}}{}
\begin{document}

\maketitle

\begin{abstract}
Automatic machine learning performs predictive modeling with high performing machine learning tools without human interference.
This is achieved by making machine learning applications parameter-free, i.e. only a dataset is provided while the complete model selection and model building process is handled internally through (often meta) optimization.
Projects like Auto-WEKA and auto-sklearn aim to solve the Combined Algorithm Selection and Hyperparameter optimization (CASH) problem resulting in huge configuration spaces.
However, for most real-world applications, the optimization over only a few different key learning algorithms can not only be sufficient, but also potentially beneficial.
The latter becomes apparent when one considers that models have to be validated, explained, deployed and maintained.
Here, less complex model are often preferred, for validation or efficiency reasons, or even a strict requirement.
Automatic gradient boosting simplifies this idea one step further, using only gradient boosting as a single learning algorithm in combination with model-based hyperparameter tuning, threshold optimization and encoding of categorical features.
We introduce this general framework as well as a concrete implementation called autoxgboost.
It is compared to current AutoML projects on 16 datasets and despite its simplicity is able to achieve comparable results on about half of the datasets as well as performing best on two.
\end{abstract}

\section{Introduction}
\label{sec:intro}

\textit{Machine Learning}, \textit{Predictive Modeling} and \textit{Artificial Intelligence} are ongoing topics in research as well as in industrial applications.
While data are gathered everywhere nowadays, many potential insights are often not fully achieved since data science and ML experts are still a rare commodity.
While many stages of a data analysis project still need to be done manually by human data scientists, model search and optimization can be done automatically.
Automatic machine learning (AutoML) simplifies the workload by making decisions for common predictive modeling tasks like regression or classification.
We distinguish between \textit{single-learner} AutoML methods which aim to make single algorithms parameter-free and more general approaches, which combine several learning algorithms into one optimization problem.
These \textit{multi-learner} methods solve the \textit{Combined Algorithm Selection and Hyperparameter optimization (CASH)} problem (\cite{autoweka}).
Modern approaches that include pre- and postpressing methods are referred to as \textit{machine learning pipeline configuration}.

There is a growing number of open source approaches for automating machine learning available for non-professionals.
As one of the first frameworks, Auto-WEKA (\cite{autoweka}) introduced a system for automatically choosing from a broad variety of learning algorithms implemented in the open source software \textit{WEKA} (\cite{weka}).
Hereby, Auto-WEKA simultaneously tunes hyperparameters over all learning algorithms model using the Bayesian optimization framework SMAC (\cite{smac}).
Similar to Auto-WEKA is auto-sklearn (\cite{autosklearn}), which is based on the scikit-learn toolkit for python and includes all of its learners as well as available preprocessing operations.
It stacks multiple models to achieve high predictive performance.
Another python-based AutoML tool is called \textit{Tree-based Pipeline Optimization Tool (TPOT)} by \cite{tpot} and uses genetic programming instead of Bayesian optimization to tune over a similar space as auto-sklearn.

Only few \textit{single-learner} AutoML methods exist.
A lot of services, for example Google's \textit{Cloud AutoML}, focus on specialized application domains like image recognition using deep neural networks.
For general machine learning tasks, \cite{probst2018hyperparameters} introduced the tuneRanger software, which automatically tunes a random forest.
Another algorithm approach is called \textit{Parameter-free STOchastic Learning (PiSTOL)} (\cite{coin}) and directly tries to optimize the generalization
performance of a learning algorithm, in a stochastic approximation way.

Our proposed approach reduces the AutoML framework to the construction of an optimal gradient boosting model (\cite{gradientboosting}), which is a strong predictive algorithm, as long as its hyperparameters are adequately tuned.
Besides tuning the hyperparameters via Bayesian optimization, categorical feature transformation is performed as a preprocessing step.
Moreover, for classification tasks, thresholds are optimized repeatedly.
By focusing on a single learning algorithm, hyperparameters can be optimized much more thoroughly and the resulting model can be analyzed and deployed much easier.


\section{Method}
\label{sec:method}

\begin{figure}[htbp]
\floatconts
  {fig: schema}
  {\caption{Workflow of the automatic gradient boosting approach. Blue lines indicate input by human.}}
  {\includegraphics[width=0.8\linewidth]{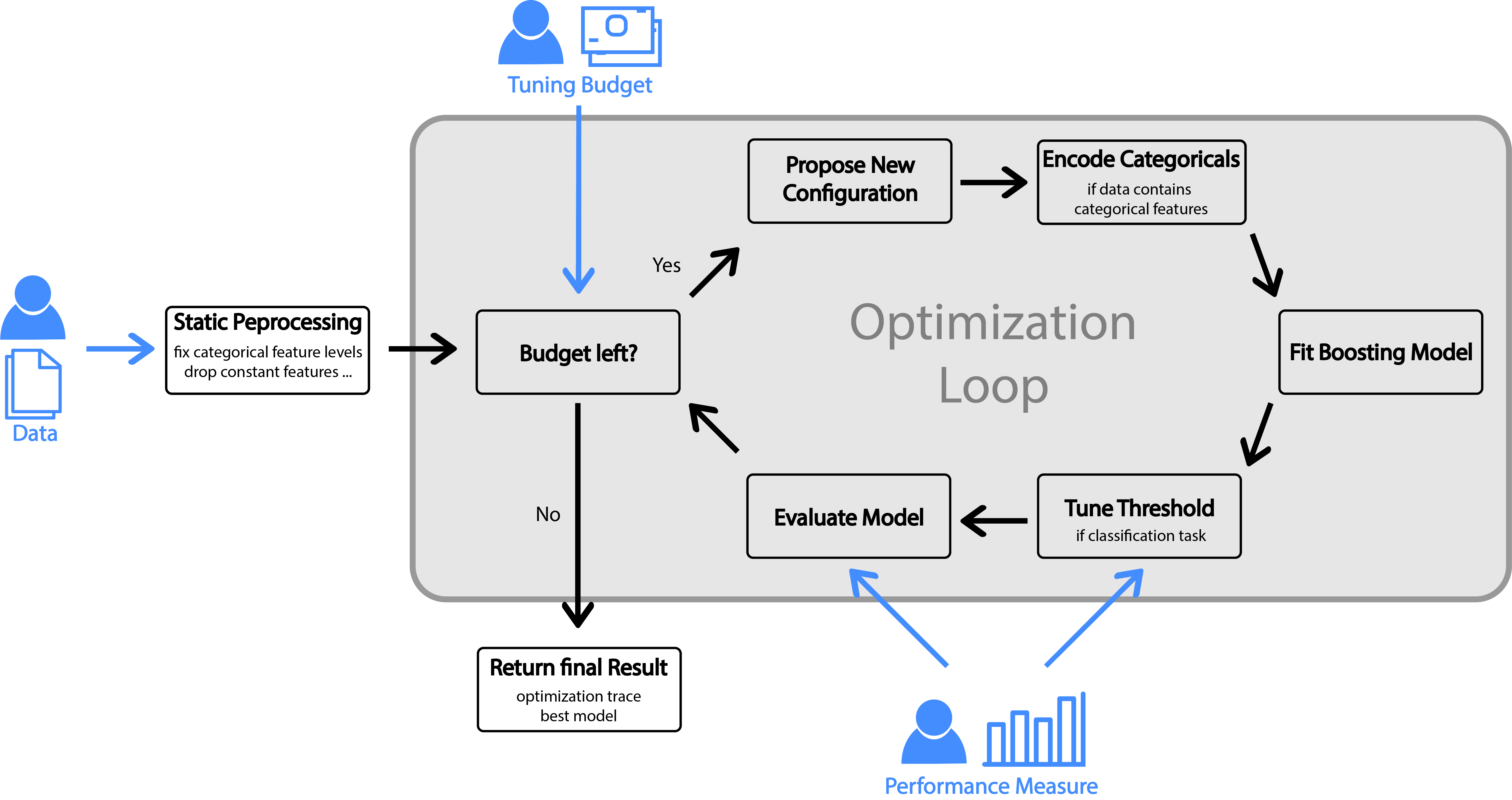}}
\end{figure}

This section introduces the structure of the automatic gradient boosting framework.
The general workflow of the approach can be seen in \figureref{fig: schema}.
Automatic gradient boosting uses gradient boosting with trees (GBT) as its only learning algorithm.
GBT is popular due to its strong predictive performance and robustness.
A large number of machine learning competitions were won by these algorithms, see \cite{xgboost} for an overview.
Furthermore it possesses multiple highly desirable properties for an AutoML system:
It is insensitive to outliers, as the trees used in gradient boosting are invariant to monotone transformations of the data, which makes scaling the data obsolete;
GBT implementations are usually able to handle missing values in the data directly by learning default split directions for missing values (\cite{xgboost});
In addition, they are capable of handling high dimensional feature spaces, as features are evaluated separately for each split in a tree, which can be parallelized and does not result in a harder optimization problem for more features.
One last important aspect to consider is that boosting can be easily adapted to tasks like ranking (\cite{li2008mcrank}) or survival analysis (\cite{chen2013gradient}).
Modern GBT frameworks like xgboost (\cite{xgboost}) or lightgbm (\cite{lightgbm}) are highly configurable with a large number of hyperparameters for regularization and optimization.
With the recent addition of dropout boosting (\cite{dart}) to these frameworks, it is possible to let the boosting algorithms behave similar, or even identical to random forests by setting hyperparameters accordingly.

The large number of hyperparameters makes tuning for GBT a necessity.
Different methods like grid or random search can be used for simple hyperparameter optimization problems, but to achieve more efficient optimization, adaptive strategies should be employed.
\textit{Sequential model-based optimization (SMBO)}, also known as \textit{Bayesian optimization} is one of the state-of-the-art adaptive hyperparameter optimization strategies (\cite{snoek2012practical}).
Depending on the difficulty of the hyperparameter space in terms of categorical and dependent hyperparameters, different surrogate models are used.
In \tableref{tab: axgb_parset} two different possible hyperparameter spaces are proposed, a fully numeric one (denoted Simple=$Y$), which can be optimized with Gaussian process surrogate models, and a more complex one (denoted Simple=$N$), which can be optimized with a random forest surrogate.
Arguably the most important hyperparameter in GBT is the number of boosting iterations, which is efficiently found by early-stopping, i.e., measuring the performance on validation data after each iteration.
An advantage of combining early stopping with SMBO is that the validation error can be directly returned to the optimizer without the need of an additional holdout set or resampling.
No necessity for internal resampling and the parallel implementation of GBT algorithms allow to mitigate the disadvantage of the sequential nature of SMBO without using parallel SMBO variants (see \cite{bischl2014moi} for an overview), such that parallel system architectures can be fully utilized.

Most boosting implementations cannot natively handle categorical variables and it is necessary to transform such features.
The simplest possibility is to encode these features into integers, with the drawback that the optimal order is unknown and a random one is used.
The concept of dummy encoding is to create one separate feature for each level, i.e., a feature $x$ with levels $a, b, c$ is dummy encoded into binary features $x^\ast_a$, $x^\ast_b$ and $x^\ast_c$.
No information is lost by this encoding but it can be infeasible for high cardinality features with a high number of unique feature levels.
A third method of transforming categorical features is \textit{impact encoding} (\citet{impact}).
Features are encoded by replacing categories with aggregated values of the target in the respective group, e.g., $\bar y|_{x=a}$ for regression or $P(y | x = a)$ for classification.
We evaluate different combinations of these encodings, mainly based on a threshold, e.g., features with less than $k$ levels are dummy encoded while integer or impact encoding is done for the remaining categorical features.
It is also possible to tune this threshold $k$ together with the GBT hyperparameters.
For datasets with few categorical features the encoding can also be learned separately for each feature.

Depending on the overall performance metric that should be optimized, it can be difficult to find the best loss function for GBT since not every performance metric can be directly plugged in as loss functions.
For binary- and multiclass classification it is often useful to optimize classification thresholds of each class directly with regard to the used performance metric.
The threshold is optimized for each iteration of the model-based optimization on the validation set that was already used for early stopping.
The resulting internal performance value is hence biased, but since the tuning error of the optimizer is biased anyways, we let this slide, especially since reducing training data size or extra resampling is much less efficient.
This design decision should be investigated in more details in future studies, though.

Combining all of the above components, we achieve a fast, scalable and robust AutoML solution that can handle categorical parameters (even with many levels), outliers and missing data, while having a much smaller configuration space compared to existing solutions.

\section{Implementation}
\label{sec:implementation}

This section introduces an implementation of the described automatic gradient boosting framework.
In general, the introduced framework could use a large number of available boosting library (e.g., xgboost or lightgbm) as well as different SMBO libraries (e.g., SMAC, Spearmint or mlrMBO).
We decided to implement it in R using xgboost as a GBT implementation, mlrMBO (\cite{mlrMBO}) for SMBO and mlr (\cite{mlr}) as a general machine learning framework as well as for threshold optimization.
For threshold optimization a multi-start linesearch is used for binary classification and for multiclass classification \textit{Generalized Simulated Annealing (GSA)} (\citet{gsa}) is applied.
The software is available via Github\footnote{\url{https://github.com/ja-thomas/autoxgboost}} and is currently able to handle binary and multiclass classification as well as regression.
It can be used in a standalone version or within the mlr framework as a learner.
Other than the data itself no further information has to be passed to autoxgboost, but it may be useful to define the performance metric (otherwise a default will be used depending on the type of the data) and the maximum runtime.
The result is a reusable machine learning pipeline based on the library mlrCPO\footnote{\url{https://github.com/mlr-org/mlrCPO}} that can be deployed or saved for later use.
Currently two different hyperparameter spaces are predefined (see \tableref{tab: axgb_parset}) and the simpler one (where Simple=$Y$) is used by default.

\begin{table}[b!]
\centering
\small
\begin{tabular}{lrrrr}
\toprule
\textbf{Name} & \textbf{Range} & \textbf{Dependency} & \textbf{$\log_2$ scale} & \textbf{Simple}\\
\midrule
  \textit{eta} & $[0.01, 0.2]$ & & N & Y\\
  \textit{gamma} & $[-7, 6]$ & & Y & Y\\
  \textit{max\_depth} & $\{3, 4, \dots, 20\}$ & & N & Y\\
  \textit{colsample\_bytree} & $[0.5, 1]$ & & N & Y\\
  \textit{colsample\_bylevel} & $[0.5, 1]$ & & N & Y\\
  \textit{lambda} & $[-10, 10]$ & & Y & Y\\
  \textit{alpha} & $[-10, 10]$ & & Y & Y\\
  \textit{subsample} & $[0.5, 1]$ & & N & Y\\
  \textit{booster} & gbtree, gblinear, dart & & N & N\\
  \textit{sample\_type} & uniform weighted & dart & N & N\\
  \textit{normalize\_type} & tree, forest & dart & N & N \\
  \textit{rate\_drop} & $[0,1]$ & dart & N & N \\
  \textit{skip\_drop} & $[0,1]$ & dart & N & N \\
  \textit{one\_drop} & TRUE, FALSE & dart & N & N \\
  \textit{grow\_policy} & depthwise, lossguide & & N & N \\
  \textit{max\_leaves} &  $\{0, 1, \dots, 8\}$ & lossguide & Y & N\\
  \textit{max\_bin}, & $\{2, 3, \dots, 9\}$ & & Y & N \\
\bottomrule
\end{tabular}
\caption{Proposed hyperparameter spaces to tune over in autoxgboost. The first $8$ parameters are defined as the simple space (default).}
\label{tab: axgb_parset}
\end{table}

\section{Benchmark}
\label{sec:bench}

We compare the performance of autoxgboost to the AutoML solutions Auto-WEKA and auto-sklearn.
In order to ensure comparability, we evaluate autoxgboost on a subset\footnote{The subset was selected to reduce computational demand. It was not cherry picked or altered in any way to improve results. A larger benchmark on more datasets is planned. An overview of the datasets can be found at \url{https://github.com/ja-thomas/autoxgboost}} of the datasets Auto-WEKA and auto-sklearn used in their respective publications.
This includes identical training- and test-data splits and the same performance measure.
The chosen datasets are very different regarding the number of numeric and factor features, as well as the number of target class levels and the train and test dataset sizes.
Hence, the datasets chosen by \citet{autoweka} serve as an adequate heterogeneous base for an initial performance evaluation in different situations.
Moreover, like in the paper of \citet{autoweka}, $25$ runs were performed.\\
The parameter settings of autoxgboost were mostly left at their default values discussed in the previous section.
However, at most $160$ tuning iterations were allowed with a maximum runtime of $10$ hours.
The benchmark was run on Intel Xeon E5-2697 v3 processors with $28$ cores and $64$gb RAM.
The hyperparameter ranges corresponded to the ones from \tableref{tab: axgb_parset} (simple).

For evaluation, $100 000$ bootstrap samples of size $4$ were drawn from all $25$ runs to simulate $4$ parallel runs.
Finally, the median of those $100 000$ mean misclassification error values is returned and presented in \tableref{tab: axgb_results}.
The bold numbers in each row indicates the best performing algorithm for the specific dataset.
We added a simple majority class baseline as an indicator that all implementations work as they should, i.e., they should significantly outperform this baseline.
\begin{table}[t!]
\centering
\footnotesize
\begin{tabular}{lrrrr}
\toprule
\textbf{Dataset}               & \textbf{baseline} & \textbf{autoxgboost} & \textbf{Auto-WEKA} & \textbf{auto-sklearn}\\
\midrule
Dexter                &       52.78  &         12.22 $$           &         7.22          &         {\textbf{5.56}}             \\
GermanCredit &          32.67 &         27.67$^*$         &         28.33     &         {\textbf{27.00}}          \\
Dorothea          &       6.09     &             {\textbf{5.22}} $$         &         6.38          &           5.51              \\
Yeast                     &       68.99 &           {\textbf{38.88}} $$       &         40.45     &         40.67           \\
Amazon              &       99.33  &       26.22 $$           &       37.56       &         {\textbf{16.00}}            \\
Secom                 &       {\textbf{7.87}}     &             {\textbf{7.87}} $$         &       {\textbf{7.87}}         &       {\textbf{7.87}}              \\
Semeion             &       92.45  &             8.38 $$          &       {\textbf{5.03}}           &           5.24             \\
Car                         &     29,15      &           1.16 $$              &       0.58          &           {\textbf{0.39}}             \\
Madelon             &     50.26    &         16.54 $$          &      21.15           &           {\textbf{12.44}}            \\
KR-vs-KP        &     48.96    &               1.67 $$        &       {\textbf{0.31}}             &           0.42            \\
Abalone               &   84.04      &         73.75$^*$        &      {\textbf{73.02}}          &           73.50       \\
Wine Quality    &     55.68    &           {\textbf{33.70}} $$         &         {\textbf{33.70}}     &           33.76         \\
Waveform        &       68.80  &       15.40$^*$               &         {\textbf{14.40}}      &           14.93         \\
Gisette                 &       50.71    &           2.48 $$          &         2.24        &             {\textbf{1.62}}           \\
Convex                &         50.00    &          22.74 $$           &         22.05     &               {\textbf{17.53}}      \\
Rot. MNIST + BI &     88.88  &            47.09$^*$           &         55.84     &             {\textbf{46.92}}      \\
\bottomrule
\end{tabular}
\caption[Autoxgboost benchmark results.]{Benchmark results are median percent error across 100 000 bootstrap samples (out of 25 runs) simulating 4 parallel runs. Bold numbers indicate best performing algorithms.
Stars indicate a relative difference of less than $5$\% to auto-sklearn.}
\label{tab: axgb_results}
\end{table}
As we can see easily in \tableref{tab: axgb_results}, only for the dataset \textit{Secom}, the baseline achieves the same performance as the AutoML frameworks.
On $9$ of the $16$ datasets, auto-sklearn provides the best results.
So does Auto-WEKA on four and autoxgboost on two datasets.
autoxgboost and Auto-WEKA slightly perform better than auto-sklearn on the \textit{Wine Quality} dataset.

\section{Conclusion}
The benchmark results of \sectionref{sec:bench} showed that autoxgboost was outperformed on the larger number of datasets by auto-sklearn, but was able to achieve competitive results on some datasets, providing state-of-the-art performance with only a single learning algorithm instead of using a whole library of possibly ensembled algorithms.
This is not too surprising as the tuning space is much smaller and on some of the datasets very different learning algorithms might have an edge.
Obviously, this is only a small initial benchmark that is not necessarily representative.
We plan to evaluate on a larger set of OpenML (\cite{vanschoren2014openml}) datasets in the future.
One clear advantage of this approach is that the resulting models are boosting models, which can be deployed more easily and allow some form of interpretability for example via feature importance and individualized feature attribution (\cite{lundberg2017consistent}).
Our AutoML implementation autoxgboost is still in an early state and some of the design decisions are not final and will be evaluated and optimized in the future.
Furthermore, we plan to extend the automatic gradient boosting framework to optimize for simultaneously sparse and well performing models using multiobjective SMBO strategies by \cite{horn2016multi}.

\newpage


\bibliography{jmlr-bib}

\begin{thebibliography}{22}
\providecommand{\natexlab}[1]{#1}
\providecommand{\url}[1]{\texttt{#1}}
\expandafter\ifx\csname urlstyle\endcsname\relax
  \providecommand{\doi}[1]{doi: #1}\else
  \providecommand{\doi}{doi: \begingroup \urlstyle{rm}\Url}\fi

\bibitem[Bischl et~al.(2014)Bischl, Wessing, Bauer, Friedrichs, and
  Weihs]{bischl2014moi}
Bernd Bischl, Simon Wessing, Nadja Bauer, Klaus Friedrichs, and Claus Weihs.
\newblock {MOI}-{MBO}: multiobjective infill for parallel model-based
  optimization.
\newblock In \emph{International Conference on Learning and Intelligent
  Optimization}, pages 173--186. Springer, 2014.

\bibitem[Bischl et~al.(2016)Bischl, Lang, Kotthoff, Schiffner, Richter,
  Studerus, Casalicchio, and Jones]{mlr}
Bernd Bischl, Michel Lang, Lars Kotthoff, Julia Schiffner, Jakob Richter, Erich
  Studerus, Giuseppe Casalicchio, and Zachary~M. Jones.
\newblock {mlr}: Machine learning in {R}.
\newblock \emph{Journal of Machine Learning Research}, 17\penalty0
  (170):\penalty0 1--5, 2016.

\bibitem[Bischl et~al.(2017)Bischl, Richter, Bossek, Horn, Thomas, and
  Lang]{mlrMBO}
Bernd Bischl, Jakob Richter, Jakob Bossek, Daniel Horn, Janek Thomas, and
  Michel Lang.
\newblock \emph{{{{mlrMBO}}: {{A Modular Framework}} for {{Model}}-{{Based
  Optimization}} of {{Expensive Black}}-{{Box Functions}}}}, 2017.

\bibitem[Chen and Guestrin(2016)]{xgboost}
Tianqi Chen and Carlos Guestrin.
\newblock {XGBoost}: A scalable tree boosting system.
\newblock In \emph{Proceedings of the 22nd ACM SIGKDD International Conference
  on Knowledge Discovery and Data Mining}, KDD '16, pages 785--794, New York,
  NY, USA, 2016. ACM.
\newblock ISBN 978-1-4503-4232-2.

\bibitem[Chen et~al.(2013)Chen, Jia, Mercola, and Xie]{chen2013gradient}
Yifei Chen, Zhenyu Jia, Dan Mercola, and Xiaohui Xie.
\newblock A gradient boosting algorithm for survival analysis via direct
  optimization of concordance index.
\newblock \emph{Computational and mathematical methods in medicine}, 2013,
  2013.

\bibitem[Feurer et~al.(2015)Feurer, Klein, Eggensperger, Springenberg, Blum,
  and Hutter]{autosklearn}
Matthias Feurer, Aaron Klein, Katharina Eggensperger, Jost Springenberg, Manuel
  Blum, and Frank Hutter.
\newblock Efficient and robust automated machine learning.
\newblock In C.~Cortes, N.~D. Lawrence, D.~D. Lee, M.~Sugiyama, and R.~Garnett,
  editors, \emph{Advances in Neural Information Processing Systems 28}, pages
  2962--2970. Curran Associates, Inc., 2015.

\bibitem[Friedman(2001)]{gradientboosting}
Jerome~H. Friedman.
\newblock Greedy function approximation: A gradient boosting machine.
\newblock \emph{Ann. Statist.}, 29\penalty0 (5):\penalty0 1189--1232, 10 2001.

\bibitem[Hall et~al.(2009)Hall, Frank, Holmes, Pfahringer, Reutemann, and
  Witten]{weka}
Mark Hall, Eibe Frank, Geoffrey Holmes, Bernhard Pfahringer, Peter Reutemann,
  and Ian~H. Witten.
\newblock The weka data mining software: An update.
\newblock \emph{SIGKDD Explor. Newsl.}, 11\penalty0 (1):\penalty0 10--18,
  November 2009.
\newblock ISSN 1931-0145.

\bibitem[Horn and Bischl(2016)]{horn2016multi}
Daniel Horn and Bernd Bischl.
\newblock Multi-objective parameter configuration of machine learning
  algorithms using model-based optimization.
\newblock In \emph{Computational Intelligence (SSCI), 2016 IEEE Symposium
  Series on}, pages 1--8. IEEE, 2016.

\bibitem[Hutter et~al.(2011)Hutter, Hoos, and Leyton-Brown]{smac}
Frank Hutter, Holger~H. Hoos, and Kevin Leyton-Brown.
\newblock \emph{Sequential Model-Based Optimization for General Algorithm
  Configuration}, pages 507--523.
\newblock Springer Berlin Heidelberg, Berlin, Heidelberg, 2011.
\newblock ISBN 978-3-642-25566-3.

\bibitem[Ke et~al.(2017)Ke, Meng, Finley, Wang, Chen, Ma, Ye, and
  Liu]{lightgbm}
Guolin Ke, Qi~Meng, Thomas Finley, Taifeng Wang, Wei Chen, Weidong Ma, Qiwei
  Ye, and Tie-Yan Liu.
\newblock Lightgbm: A highly efficient gradient boosting decision tree.
\newblock In I.~Guyon, U.~V. Luxburg, S.~Bengio, H.~Wallach, R.~Fergus,
  S.~Vishwanathan, and R.~Garnett, editors, \emph{Advances in Neural
  Information Processing Systems 30}, pages 3149--3157. Curran Associates,
  Inc., 2017.

\bibitem[Li et~al.(2008)Li, Wu, and Burges]{li2008mcrank}
Ping Li, Qiang Wu, and Christopher~J Burges.
\newblock Mcrank: Learning to rank using multiple classification and gradient
  boosting.
\newblock In \emph{Advances in neural information processing systems}, pages
  897--904, 2008.

\bibitem[Lundberg and Lee(2017)]{lundberg2017consistent}
Scott~M Lundberg and Su-In Lee.
\newblock Consistent feature attribution for tree ensembles.
\newblock \emph{arXiv preprint arXiv:1706.06060}, 2017.

\bibitem[Micci-Barreca(2001)]{impact}
Daniele Micci-Barreca.
\newblock A preprocessing scheme for high-cardinality categorical attributes in
  classification and prediction problems.
\newblock \emph{SIGKDD Explor. Newsl.}, 3\penalty0 (1):\penalty0 27--32, July
  2001.
\newblock ISSN 1931-0145.

\bibitem[Olson et~al.(2016)Olson, Urbanowicz, Andrews, Lavender, Kidd, and
  Moore]{tpot}
Randal~S. Olson, Ryan~J. Urbanowicz, Peter~C. Andrews, Nicole~A. Lavender,
  La~Creis Kidd, and Jason~H. Moore.
\newblock Automating biomedical data science through tree-based pipeline
  optimization.
\newblock In Giovanni Squillero and Paolo Burelli, editors, \emph{Applications
  of Evolutionary Computation}, pages 123--137, Cham, 2016. Springer
  International Publishing.
\newblock ISBN 978-3-319-31204-0.

\bibitem[Orabona(2014)]{coin}
Francesco Orabona.
\newblock Simultaneous model selection and optimization through parameter-free
  stochastic learning.
\newblock In \emph{Advances in Neural Information Processing Systems}, pages
  1116--1124, 2014.

\bibitem[Probst et~al.(2018)Probst, Wright, and
  Boulesteix]{probst2018hyperparameters}
Philipp Probst, Marvin Wright, and Anne-Laure Boulesteix.
\newblock Hyperparameters and tuning strategies for random forest.
\newblock \emph{arXiv preprint arXiv:1804.03515}, 2018.

\bibitem[Rashmi and Gilad-Bachrach(2015)]{dart}
KV~Rashmi and Ran Gilad-Bachrach.
\newblock Dart: Dropouts meet multiple additive regression trees.
\newblock In \emph{International Conference on Artificial Intelligence and
  Statistics}, pages 489--497, 2015.

\bibitem[Snoek et~al.(2012)Snoek, Larochelle, and Adams]{snoek2012practical}
Jasper Snoek, Hugo Larochelle, and Ryan~P Adams.
\newblock Practical bayesian optimization of machine learning algorithms.
\newblock In \emph{Advances in neural information processing systems}, pages
  2951--2959, 2012.

\bibitem[Thornton et~al.(2013)Thornton, Hutter, Hoos, and
  Leyton-Brown]{autoweka}
C.~Thornton, F.~Hutter, H.~H. Hoos, and K.~Leyton-Brown.
\newblock Auto-{WEKA}: Combined selection and hyperparameter optimization of
  classification algorithms.
\newblock In \emph{Proc.~of KDD-2013}, pages 847--855, 2013.

\bibitem[Tsallis and Stariolo(1996)]{gsa}
Constantino Tsallis and Daniel~A. Stariolo.
\newblock Generalized simulated annealing.
\newblock \emph{Physica A: Statistical Mechanics and its Applications},
  233\penalty0 (1):\penalty0 395 -- 406, 1996.
\newblock ISSN 0378-4371.

\bibitem[Vanschoren et~al.(2014)Vanschoren, Van~Rijn, Bischl, and
  Torgo]{vanschoren2014openml}
Joaquin Vanschoren, Jan~N Van~Rijn, Bernd Bischl, and Luis Torgo.
\newblock Openml: networked science in machine learning.
\newblock \emph{ACM SIGKDD Explorations Newsletter}, 15\penalty0 (2):\penalty0
  49--60, 2014.

\end{thebibliography}

\end{document}